\pdfoutput=1
\documentclass{article}

\PassOptionsToPackage{numbers, compress}{natbib}
\usepackage[dblblindworkshop,final]{neurips_2026}
\workshoptitle{Foundation Models for the Brain and Body}
\usepackage[utf8]{inputenc}
\usepackage[T1]{fontenc}
\usepackage{hyperref}
\usepackage{url}
\providecommand{\doi}[1]{\href{https://doi.org/#1}{doi: \nolinkurl{#1}}}
\usepackage{booktabs}
\usepackage{tabularx}
\usepackage{multirow}
\usepackage{placeins}
\usepackage{amsfonts,amsmath,amssymb,mathtools,bm,nicefrac,microtype,xcolor}
\usepackage{graphicx}
\usepackage{etoolbox}
\AtBeginEnvironment{figure}{\small}

\newcommand{\C}{\mathbb{C}}
\newcommand{\R}{\mathbb{R}}
\newcommand{\by}{\mathbf{y}}
\newcommand{\bz}{\mathbf{z}}
\newcommand{\bgamma}{\bm{\gamma}}
\newcommand{\bomega}{\bm{\omega}}

\title{Self-Supervised Learning with Recursive Quadrature Filters for Neural Time Series}

\author{Mirko Morello$^{1, *}$ \hspace{14pt} Shivang Rawat$^{1}$ \hspace{14pt} David J. Heeger$^{1, 2}$ \vspace{10pt} \\
$^1$ Telepath \\
$^2$ New York University\\
\\
$^*$\texttt{mirko@telepath.com}
}

\begin{document}

\maketitle

\begin{abstract}
We develop self-supervised methods for training Recursive Quadrature
Filter (RQF) encoders on unlabeled time series. RQFs are biologically motivated,
complex-valued temporal filters that form a special class of diagonal
state-space models (SSMs).
Each RQF acts as a band-pass filter whose selectivity is determined
by two learnable parameters: its tuning frequency and bandwidth.
We separately introduce a self-supervised learning objective based on prediction error; 
the encoder is trained so that its current
state predicts its own future activity. In the \textit{windowed} mode, the
summary of the current window predicts the summary of a later window; in
the \textit{dense} mode, the state at the current timestep predicts
future log-power. Each RQF encoder
is pretrained on a dataset and then a readout (e.g., a task-specific classifier) is trained on the output of the RQF encoder, with the encoder either frozen or fully fine-tuned. 
For multichannel EEG and intracranial recordings, a
shared encoder processes each channel separately, and a spatial readout
combines the channel representations using electrode locations. 
Across four neural benchmarks using scalp or intracranial EEG, the pretrained RQF system
performs close to the state of the art while using a small fraction of
the parameters of the competing models. 
For example, a 110k-parameter RQF system achieves
$0.9247 \pm 0.0022$ test AUROC on 100-s TUAB segments, above SAMBA
(0.9054) at the same context with 9 times as many parameters, and
$0.9040 \pm 0.0019$ on 10-s segments, comparable to EEG
foundation models with 3.2M--369M parameters.
\end{abstract}

\section{Introduction}
\label{sec:intro}

Self-supervised pretraining has become the default route to
general-purpose representations of physiological time series. The
leading EEG encoders often follow the recipe established in language and
vision: a learned tokenizer maps raw signal into patches or discrete
codes, a Transformer is trained to predict masked patches or codes, and
the encoder is scaled to millions or hundreds of millions of
parameters~\cite{jiang2024labram,wang2025cbramod,reve2025}.
The resulting systems perform well across benchmarks. However,
because scale, learned input embeddings, and masked pretraining arrive
together, it is difficult to tell how much of the result comes from
capacity and how much from the recipe, and the encoder does not explicitly
parameterize oscillatory dynamics. This leaves open whether a
compact encoder whose dynamics match the 
statistical regularities of the input signals can achieve the same
performance range without either scale or masking.

Here, we develop self-supervised learning
with Recursive Quadrature Filters (RQFs), 
a class of biologically motivated complex-valued temporal
filters~\cite{rqf2026,heeger2020recurrent}. Each RQF learns a tuning
frequency and a bandwidth and behaves as a band-pass filter. A stack of
RQFs is therefore inherently suited to physiological signals such as EEG
and intracranial recordings, whose information is carried by rhythms in
distinct frequency bands. 
RQFs are \textit{prospective} by
construction; the recurrence evolves under a phase-rotation prior that
extrapolates the oscillation it currently tracks
(Section~\ref{sec:background}). 
In addition, the bottom-up input to each layer is
itself prospective; the layer is driven by a first-order look-ahead of
its input~\cite{rqf2026}. 
Both the phase-rotation prior and the prospective input 
enable the encoder to predict its immediate future, such that prediction error is a natural learning objective.
Because RQFs are a special case of diagonal state-space models
(SSMs), they can be trained with the same computationally efficient parallel-scan algorithms as diagonal SSMs.

An RQF encoder learns by predicting its own future activity, i.e., based on statistical regularities in the input signals, with no task labels.
The self-supervised objective is a loss on the encoder parameters that compares an earlier
representation with one produced far later, at horizons of tens of
milliseconds to seconds. 
A prediction head maps an earlier representation to
the representation that the same encoder produces later. 
The \textit{windowed} objective divides the state sequence into windows, summarizes each with parameter-free or learned pooling, and predicts the summary of a later window,
whereas in the \textit{dense} objective the state at each timestep predicts future log-power.
The input-side prospective correction is a fixed first-order look-ahead of
each layer's bottom-up drive that compensates for integration lag without
adding learnable parameters~\cite{rqf2026}. The self-supervised objective
trains the encoder, including frequencies and bandwidths of each RQF,
together with the prediction head to predict future activity.
Appendix~\ref{app:related} relates this objective to CPC, JEPA, and
previous self-supervised state-space encoders.

To assess performance of RQFs with self-supervised prediction-error based learning, 
we designed experiments on four benchmarks that separate end-to-end system performance from
representation quality. After pretraining, the encoder is reused for a
supervised task either frozen or fully fine-tuned, under audited splits
and validation-only model selection (Section~\ref{sec:setup}).
1) TUAB tests
whether the complete windowed system, pretrained only on the target
corpus, performs within the foundation-model range, and a paired
label-budget sweep measures when pretraining is most beneficial.
2) BrainTreebank compares the
frozen and fully fine-tuned encoder.
3) ISRUC-SG1 tests the windowed system on five-class sleep
staging~\cite{khalighi2016isruc}.
4) BCI-IV-2a tests cross-corpus transfer
by using TUEG-pretrained weights for motor imagery, i.e., 
the pretraining and downstream corpora differ.

We find that small encoders perform surprisingly well. 
Across all four benchmarks the pretrained RQF system performs close to
the state of the art with a small fraction of the parameters of the
competing models (Figure~\ref{fig:pareto}). 
For example, with 110k
parameters, windowed pretraining followed by full fine-tuning achieves
$0.9247 \pm 0.0022$ test AUROC on 100-s TUAB segments, above SAMBA at
the same context (0.9054, 1.0M parameters), and $0.9040 \pm 0.0019$ on
10-s segments, comparable to EEG foundation models with
3.2M--369M parameters.
On intracranial BrainTreebank, we achieve
$0.8822 \pm 0.0011$ AUROC with the pretrained RQF encoder frozen,
2.9 points above the best published model (BaRISTA) that has 3.4 times more parameters.

\section{Background: Recursive Quadrature Filters}
\label{sec:background}

A \textit{Recursive Quadrature Filter} (RQF)~\cite{rqf2026} is a scalar
complex-valued temporal filter derived from an energy-based model 
of neural circuit function~\citep{heeger2017theory}. 
It receives an input drive $z(t) \in \C$, and its
state $y(t) \in \C$ evolves as
\begin{equation}
    \tau\,\dot y \;=\; -y + \gamma\,z + (1-\gamma)(1 + \iota\omega\tau)\,y,
    \label{eq:rqf_scalar}
\end{equation}
where $\tau > 0$ is a base time constant, fixed during pretraining
(Appendix~\ref{app:enablement}); $\omega > 0$ is the
learnable frequency parameter; and $\gamma \in (0,1)$ is the learnable
bandwidth parameter. With $\alpha \triangleq \gamma/\tau$,
$\omega_0 \triangleq (1-\gamma)\omega$, and
$\lambda \triangleq -\alpha + \iota\omega_0$, Eq.~\eqref{eq:rqf_scalar}
collapses to $\dot y = \lambda y + \alpha z$, a band-pass filter with unit
peak gain at $\omega_0$ and half-power bandwidth $2\alpha$ that is stable
for any $\gamma \in (0,1)$. The prior term $(1+\iota\omega\tau)\,y$
extrapolates the state by one time constant to first order under a
phase-rotation model, making the recurrence prospective by construction.
The complex state makes each scalar filter oscillatory. 
The state is complex-valued even when the input $z$ is real, with its real and imaginary parts representing the output of a quadrature pair of causal, temporal bandpass filters (i.e., with the same frequency response amplitude but with a $90^\circ$ phase shift at each frequency).
Hence, the modulus of the output $|y|$
follows the amplitude envelope of the filter's band
regardless of phase. In real coordinates, each RQF is a damped harmonic
oscillator~\cite{rqf2026}. The phase-rotation prior gives the recurrence
a model of how an oscillation continues, which a prediction objective
can exploit.

An encoder stacks $L$ layers of $d$ such filters with learned 
parameters $\bomega^{(l)}, \bgamma^{(l)} \in \R^{d}$. The input to
layer $l>1$ is $\bz_n^{(l)} = \mathbf{W}^{(l)}\rho\!\left(\by_n^{(l-1)}\right)$,
where $\mathbf{W}^{(l)}$ is a complex inter-layer mixing matrix and $\rho$
is an activation function (also called an output nonlinearity),
e.g., split-ReLU applied to the real and imaginary components. The first
layer receives the waveform at the dataset-specific sampling rate.
In addition, we make the
input to each layer prospective~\cite{rqf2026}: each layer is driven by $\bz^{(l)} + \tau\,\dot{\bz}^{(l)}$
in place of $\bz^{(l)}$, a first-order look-ahead of its input. Under
zero-order-hold discretization this becomes a two-tap update on the
current and preceding input samples. These taps introduce no additional
trainable parameters and leave the recurrent transition unchanged.
Self-supervised pretraining optimizes the encoder, including its
phase-prior frequencies and bandwidths, together with the prediction head
to predict activity at longer horizons, from tens of milliseconds to
seconds (Section~\ref{sec:method}).

The top-layer
states $\by_n^{(L)} \in \C^{d}$ form the representation used during
pretraining and by the downstream task readout. Because the RQF is a
special case of a diagonal complex state-space model, the encoder trains
with the same parallel-scan algorithms as modern SSMs. The encoders 
that we trained for the results reported below have 30k to 140k parameters
(Appendix~\ref{app:enablement}).

\section{Self-supervised learning with RQF encoders}
\label{sec:method}

The RQF encoder maps a raw sequence of $T$ input
samples to top-layer states $\by_1^{(L)},\ldots,\by_T^{(L)}$ with
$\by_n^{(L)} \in \C^{d}$. Self-supervised pretraining uses no task labels: the encoder
is trained to predict its own future states from its current ones. 
We have tested two learning objectives.
The \textit{windowed} objective predicts between
summaries of the state sequence, whereas in the \textit{dense} objective
the state at each timestep predicts future log-power. For both objectives, 
a source representation $\mathbf{s}_a$ is read
from the states corresponding to an earlier time $a$; a learned \textit{prediction
head} $f_\phi$ maps it to an estimate $\hat{\mathbf{t}}_a = f_\phi(\mathbf{s}_a)$
of a target $\mathbf{t}^{+}_a$ corresponding to a later time; and the estimate
is scored by cosine similarity, after $\ell_2$ normalization, against the
true target and against $K$ negatives $\mathbf{t}^{-}_{a,k}$, e.g., 
corresponding to other times. 
With $\mathcal{A}$ the set of sources in a batch and
$\beta$ an inverse temperature, the loss is the InfoNCE objective~\cite{oord2018cpc}:
\begin{equation}
    \mathcal{L}_{\mathrm{NCE}} \;=\; -\,\frac{1}{|\mathcal{A}|}
    \sum_{a \in \mathcal{A}}
    \log \frac{\exp(\beta \operatorname{sim}^{+}_{a})}
    {\exp(\beta \operatorname{sim}^{+}_{a})
    + \sum_{k=1}^{K} \exp(\beta \operatorname{sim}^{-}_{a,k})},
    \label{eq:infonce}
\end{equation}
where $\operatorname{sim}^{+}_{a}$ is the cosine similarity between
$\hat{\mathbf{t}}_a$ and $\mathbf{t}^{+}_a$; $\operatorname{sim}^{-}_{a,k}$
is the cosine similarity between $\hat{\mathbf{t}}_a$ and the $k$th
negative; and targets are detached. 
The prediction head is discarded after
pretraining. The two objectives differ in the representations that are being used for prediction ($\mathbf{s}_a$ and
$\mathbf{t}^{+}_a$), either temporal samples or abstracted summaries of RQF outputs, and in the temporal scale of the prediction horizon.

\paragraph{Windowed future prediction.}
The windowed objective predicts how the representation evolves at a
coarser temporal scale. We divide the top-layer states into fixed,
non-overlapping windows and summarize each one into a vector
$\mathbf{t}_j$: the speech configuration uses a parameter-free mean,
whereas the multichannel configurations use a learned pooler that
combines the mean real and imaginary parts of RQF outputs in the window with its
final state, followed by a linear projection and layer normalization.
The source is the summary of window $j$ and the target is the summary of
window $j+h_w$, where the horizon $h_w$ is an integer number of windows. Negatives are other
window summaries in the batch, from the same input or from other
inputs. A regression configuration instead minimizes the mean-squared
error between the estimated and true future summaries and needs no
negatives; the BrainTreebank diagnostic in Appendix~\ref{app:additional-results} uses
this regression loss.

\paragraph{Dense per-timestep prediction.}
In the dense objective, the state at the current timestep predicts future log-power. The source is the real and imaginary
components of the state $\by_n^{(L)}$ at an anchor timestep $n$. The
target is built from the states in a future interval of $N_w \geq 1$
samples that begins $\Delta \geq 1$ samples after the anchor, so
$\Delta$ is the prediction horizon in samples. Specifically, the target
is \textit{future log-power}, capitalizing on the quadrature pair of filters to compute the demodulated mean power of the
RQF outputs:
\begin{equation}
    \mathbf{g}_n \;=\; \log\!\Big(\epsilon \,+\, \tfrac{1}{N_w}
    \textstyle\sum_{u=n+\Delta}^{n+\Delta+N_w-1} \big|\by_u^{(L)}\big|^2\Big),
    \label{eq:dense_logpower}
\end{equation}
where $u$ indexes temporal samples, with $\epsilon$ a small constant. 
Negatives are the log-power targets of other anchors. 

\begin{figure}[t]
\centering
\includegraphics[width=0.64\textwidth]{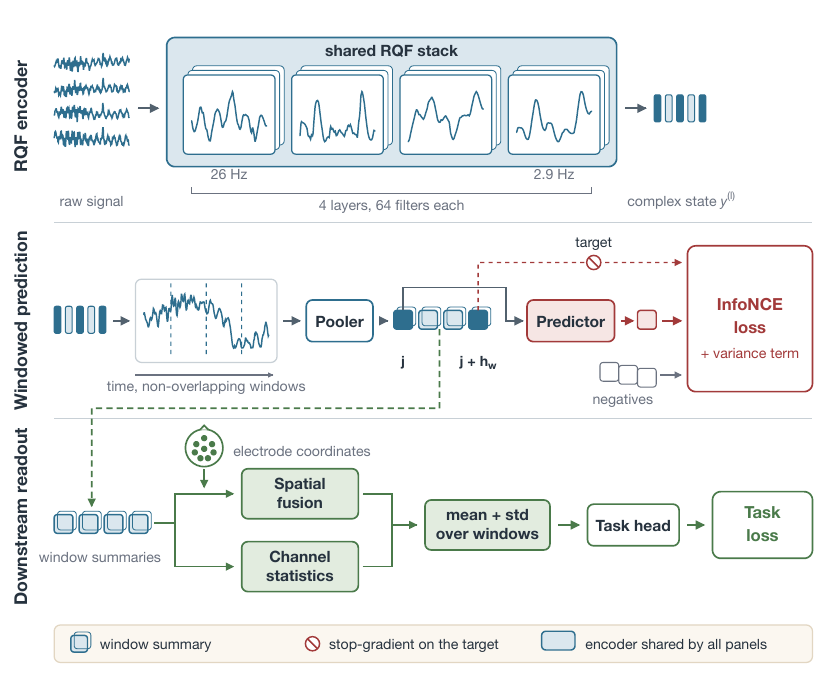}
\caption{\textbf{Windowed self-supervised training and the multichannel
task readout.} The three panels share one encoder. \textbf{Top:} A single
RQF stack encodes every channel independently 
with four layers of 64 filters; the front plot of each
layer shows the state magnitude of one filter, driven by a synthetic
EEG-like probe input. The initialization ceiling
of the frequency parameters halves from layer to layer, and the plotted
exemplars run from 26 to 2.9\,Hz. \textbf{Middle:} The state sequence is
cut into non-overlapping windows, the learned pooler maps the mean and
the real and imaginary parts of the final state of each window to a
summary, and the predictor maps the summary of window $j$ to the summary
of window $j+h_w$. That later summary is the positive
and enters with a
stop-gradient, the negatives are other window summaries in the batch, and
a variance term accompanies the InfoNCE objective. \textbf{Bottom:} The
same summaries take two routes that are concatenated. One projects them
and fuses them across channels (e.g., using the electrode
coordinates), and the other takes their mean, standard deviation, and
maximum across channels. The concatenation is pooled by its mean and
standard deviation over windows before the task head (e.g., task-specific classifier). Dimensions, horizons, and optimization settings
are in Table~\ref{tab:hyperparams}.}
\label{fig:architecture}
\end{figure}

\paragraph{Multichannel composition.}
Multichannel recordings add a spatial component without altering the
temporal prediction task. During pretraining, the shared encoder
processes each channel of each input as an independent single-channel
sequence, and the pretext task receives neither electrode identity nor
position. Spatial information enters only downstream, where a
coordinate-aware readout combines the per-channel window summaries using
the electrode coordinates. Figure~\ref{fig:architecture}
illustrates the windowed path and this readout.
Appendix~\ref{app:enablement} gives the pooler and readout construction,
anchor sampling, target calibration, the per-configuration negative
exclusion policies, and the dataset-specific windows, horizons, and loss
constants.

\section{Evaluation design}
\label{sec:setup}

The benchmarks address three questions: 1) whether the complete RQF system
can match the performance range of much larger encoders; 2) when
pretraining benefits a fixed downstream architecture; and 3) how temporal
aggregation affects performance.
TUAB tests performance with different amounts of labeled data, using
patient-disjoint splits following BIOT~\cite{yang2023biot}. Details of the
test set are provided in Appendix~\ref{app:additional-results}.
BrainTreebank (BTB) tests whether the pretrained encoder supports sentence-onset
detection~\cite{wang2024braintreebank}; we pretrain on BaRISTA's
17-session partition and evaluate using chronological splits within its
seven held-out sessions~\cite{oganesian2025barista}. ISRUC-SG1 tests the
windowed learning objective on five-class sleep staging~\cite{khalighi2016isruc}.
BCI-IV-2a tests
cross-corpus transfer by pretraining the encoder
on a subset of the TUH EEG corpus (TUEG)
and testing with four-class motor imagery~\cite{tangermann2012bcic}.
Throughout, \textit{transfer} refers to reusing the pretrained encoder
weights for a supervised downstream task after discarding the prediction
head. Except for BCI-IV-2a, the pretraining and downstream corpora
coincide, so transfer is across tasks within a corpus rather than across
corpora; we write \textit{cross-corpus transfer} for the latter.

We compare each pretrained encoder with a control that starts from the
same initial weights but skips pretraining (\textit{paired init.}). 
During supervised training, the encoder is either kept fixed
(\textit{frozen}) or updated (\textit{fully fine-tuned}, full-FT).
For each benchmark, we use the test set only for final
evaluation. 

See Appendices~\ref{app:additional-results} and~\ref{app:enablement} 
for full evaluation details including definitions of the performance metrics, 
and for a complete report of all of the results including
supplementary results on a speech benchmark.

\section{Results}
\label{sec:results}

\begin{figure}[t]
\centering
\includegraphics[width=0.92\textwidth]{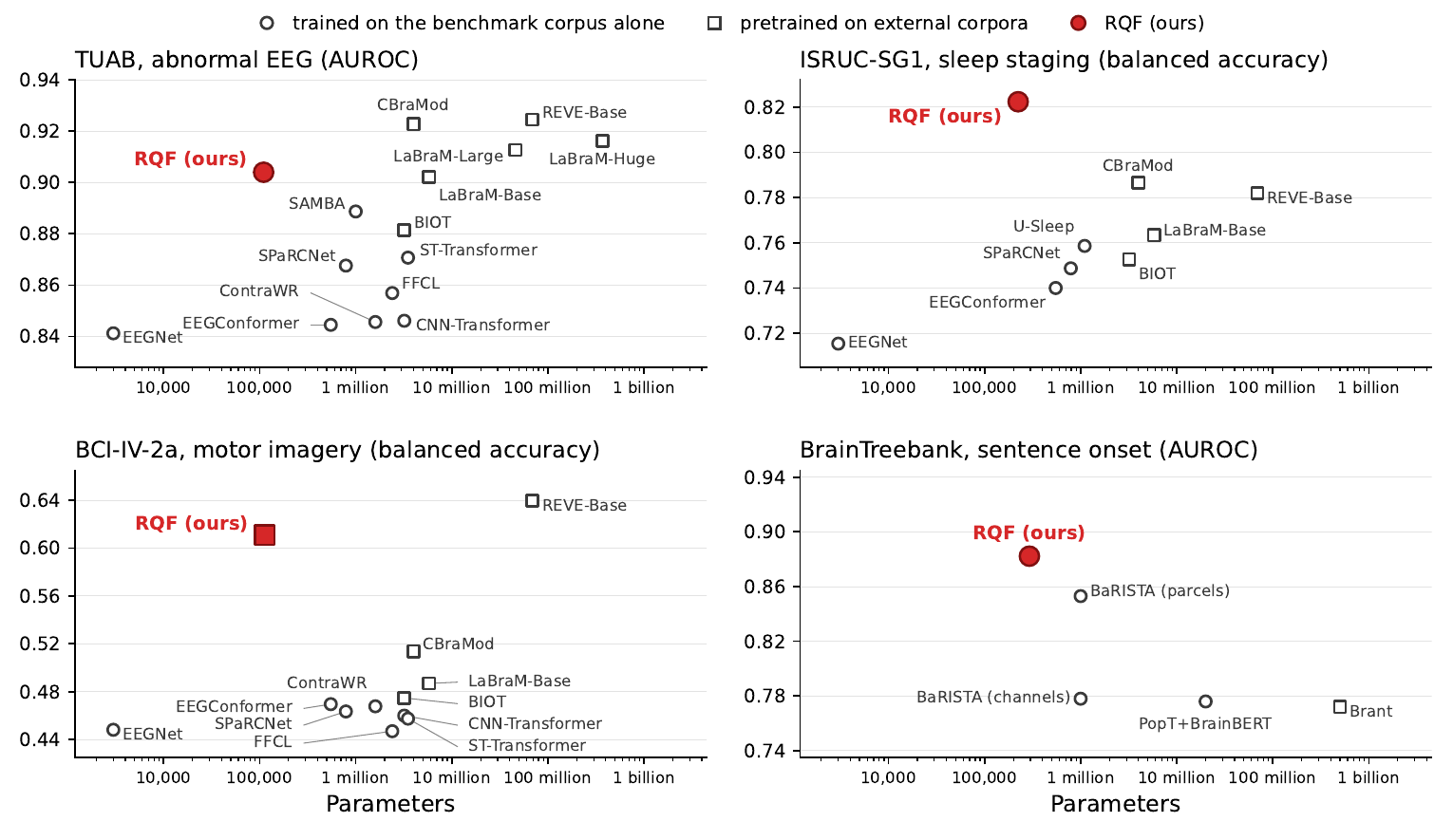}
\caption{\textbf{Test performance versus parameter count on four benchmarks.}
Each panel uses the metric named in its title. Red marks our strongest
pretrained RQF configuration; circles denote training on the benchmark corpus
and squares denote external-corpus pretraining. Published values are transcribed from
their source papers, whose protocols differ from ours
(Appendix~\ref{app:datasets}), so the panels give context, not
controlled comparisons. 
All TUAB points, ours included, use 10-s segments; our 100-s result is
compared with SAMBA at the same context in the text and Table~\ref{tab:tuab100}.
All BrainTreebank points use BaRISTA's chronological-split
evaluation (its Appendix Table~16)~\cite{oganesian2025barista}. 
Full values are in Appendix~\ref{app:additional-results}.}
\vspace{-12pt}
\label{fig:pareto}
\end{figure}

Our results are of two kinds. Controlled comparisons fix the data,
preprocessing, splits, and training settings and vary one factor
(pretraining, freezing, objective, or label budget). Comparisons with
published models use values from their source papers; these results give
context but do not isolate a cause because the published protocols may differ from ours (Appendix~\ref{app:datasets}). When comparing model sizes, we count the real and imaginary parts of the complex-valued RQF parameters as 2 separate parameters.

\paragraph{A compact RQF compares favorably with large-scale models.}
On 100-s TUAB segments, windowed pretraining followed by full
fine-tuning achieves $0.9247 \pm 0.0022$ test AUROC with 110{,}002
parameters, above SAMBA at the same context (0.9054, 9 times
as many parameters). On the 10-s segments of the other published models,
which Figure~\ref{fig:pareto} plots,
it achieves $0.9040 \pm 0.0019$, above LaBraM-Base (0.9022, 53 times as
many). LaBraM-Large, LaBraM-Huge, CBraMod, and REVE-Base outperform it
by 0.9, 1.2, 1.9, and 2.0 points with 418, 3{,}355, 36, and 627 times as
many parameters
(Figure~\ref{fig:pareto}; Tables~\ref{tab:tuab} and~\ref{tab:tuab100}).

Likewise, for BCI-IV-2a, full
fine-tuning of TUEG-pretrained weights with Euclidean alignment achieves
$0.6108 \pm 0.0089$ balanced accuracy, 9.7 points above CBraMod, which
uses 35 times as many parameters, and 2.9 points below REVE-Base, which
uses 612 times as many (Figure~\ref{fig:pareto} and Appendix Table~\ref{tab:bcic}).
With this preprocessing, our RQF trained without SSL performs within
0.9 balanced-accuracy points of the versions pretrained on BCI-IV-2a or TUEG.

For ISRUC-SG1, the RQF system achieves $0.8223 \pm 0.0028$ balanced
accuracy with 223{,}733 parameters, including the context
module and classifier. It outperforms all of the comparison models,
including REVE-Base, which uses 308 times as many parameters
(Figure~\ref{fig:pareto} and Appendix Table~\ref{tab:isruc}).
Its paired initialization achieves $0.8252 \pm 0.0016$, so pretraining
adds no measurable gain for this benchmark (95\% interval over test subjects, $-1.1$
to $0.6$ points). Both use context from 20 consecutive 30-second
segments to predict the sleep stage of each.

\paragraph{Under full fine-tuning, pretraining helps most when labels are
scarce.}
We use the TUAB benchmark to determine the benefit of pretraining
with limited labeled data. We start two copies of the same encoder, both
initialized with the same random weights, and we pretrain one of them
with the windowed objective. We then add the readout and
train both networks  
on the same labeled subset, allowing all weights (encoder + readout) to
change. With only 1\% of the training labels, the pretrained model achieves
$0.7001 \pm 0.0722$ AUROC on 10-s segments, compared to
$0.6509 \pm 0.0230$ without pretraining, a gain of 4.9 points
(95\% interval over test patients, 1.4 to 8.3), although one of three
seeds is worse with pretraining. With all labels available, the two
are level (0.9040 and 0.9046). Thus, self-supervised pretraining is most
beneficial here when the supervised task provides little training data
(Appendix Table~\ref{tab:tuab_label_efficiency}).

\paragraph{The prediction objective shapes the learned representation.}
We use BrainTreebank to compare the two objectives.
After pretraining, we keep the encoder frozen and train a task-specific
readout. This tests whether the learned
representation already contains information useful for the task.
Windowed prediction (predicting window summaries) gives $0.8822 \pm 0.0011$
AUROC over five downstream seeds, 5.7 points above the same encoder
frozen at paired initialization ($0.8247 \pm 0.0064$; 95\% interval over
sessions, 2.8 to 9.4 points) and 1.5 points above dense prediction, which
predicts future log-power from individual temporal samples ($0.8670 \pm 0.0055$;
0.6 to 2.6 points; Appendix Table~\ref{tab:btb}).

Updating the encoder when training the task-specific readout does not necessarily improve
performance over keeping the encoder frozen. After windowed pretraining, the frozen
encoder scores 1.2 points above full fine-tuning (0.8822 and 0.8706; 0.2 to 2.3 points),
while paired initialization orders the other way (0.8247 frozen and 0.8459 fully
fine-tuned; 0.4 to 4.1 points).
These results suggest that
on BrainTreebank, freezing the pretrained representation can be beneficial
even when there is enough labeled data to update all weights.

\section{Discussion, limitations, and conclusion}
\label{sec:discussion}

Three findings follow from the experiments. 1) A compact RQF encoder
whose units learn a tuning frequency and a bandwidth is a viable
alternative to Transformer-based models 
when the statistical regularities in the input signals are characterized
by oscillations. On TUAB, for example, the fine-tuned model relies most on
a few filters tuned to the alpha and beta bands (Appendix Table~\ref{tab:band-ablation}).
2) The choice of source and target matters: predicting the summary of a later window from the summary of the current one yields stronger frozen representations than predicting future log-power from the state at the current timestep, although both lie well above the paired-initialization control. 3) Pretraining
acts as a prior whose value depends on how the representation is used. The gain with
pretraining is greatest when labels are scarce or the encoder is frozen.
In our experiments, the gains are smaller under full fine-tuning, where the downstream loss is free to
reshape the encoder.

The scope is limited in two ways. 1) Apart from the cross-corpus
TUEG-to-BCI-IV-2a result, each corpus has its own pretrained weights, target
geometry, and readout. 2) The comparisons with foundation models do not
isolate the oscillatory bias from tokenization, masking, or scale.

In summary, future-state prediction is a practical way to
pretrain RQF encoders on scalp EEG and intracranial recordings 
(also speech, see Appendix~\ref{app:additional-results}).
A small
recurrent encoder learns competitive representations when its dynamics
and its objective share a model of time, and RQFs provide that model
for oscillatory signals.

\bibliographystyle{unsrtnat}
\bibliography{references}

\begin{thebibliography}{35}
\providecommand{\natexlab}[1]{#1}
\providecommand{\url}[1]{\texttt{#1}}
\expandafter\ifx\csname urlstyle\endcsname\relax
  \providecommand{\doi}[1]{doi: #1}\else
  \providecommand{\doi}{doi: \begingroup \urlstyle{rm}\Url}\fi

\bibitem[Jiang et~al.(2024)Jiang, Zhao, and Lu]{jiang2024labram}
Wei-Bang Jiang, Li-Ming Zhao, and Bao-Liang Lu.
\newblock Large brain model for learning generic representations with
  tremendous {EEG} data in {BCI}.
\newblock In \emph{International Conference on Learning Representations}, 2024.
\newblock URL \url{https://arxiv.org/abs/2405.18765}.

\bibitem[Wang et~al.(2025)Wang, Zhao, Luo, Zhou, Jiang, Li, Li, and
  Pan]{wang2025cbramod}
Jiquan Wang, Sha Zhao, Zhiling Luo, Yangxuan Zhou, Haiteng Jiang, Shijian Li,
  Tao Li, and Gang Pan.
\newblock {CBraMod}: A criss-cross brain foundation model for {EEG} decoding.
\newblock In \emph{International Conference on Learning Representations}, 2025.
\newblock URL \url{https://arxiv.org/abs/2412.07236}.

\bibitem[El~Ouahidi et~al.(2025)El~Ouahidi, Lys, Th{\"o}lke, Farrugia,
  Pasdeloup, Gripon, Jerbi, and Lioi]{reve2025}
Yassine El~Ouahidi, Jonathan Lys, Philipp Th{\"o}lke, Nicolas Farrugia, Bastien
  Pasdeloup, Vincent Gripon, Karim Jerbi, and Giulia Lioi.
\newblock {REVE}: A foundation model for {EEG} --- adapting to any setup with
  large-scale pretraining on 25,000 subjects.
\newblock \emph{Advances in Neural Information Processing Systems},
  38:\penalty0 25583--25619, 2025.
\newblock \doi{10.52202/085713-0760}.
\newblock arXiv:2510.21585.

\bibitem[Rawat et~al.(2026)Rawat, Morello, Morone, and Heeger]{rqf2026}
Shivang Rawat, Mirko Morello, Flaviano Morone, and David~J. Heeger.
\newblock Prospective coding improves learning in deep continuous-time
  recurrent networks.
\newblock \emph{arXiv preprint arXiv:2609.04134}, 2026.
\newblock URL \url{https://arxiv.org/abs/2609.04134}.

\bibitem[Heeger and Zemlianova(2020)]{heeger2020recurrent}
David~J Heeger and Klavdia~O Zemlianova.
\newblock A recurrent circuit implements normalization, simulating the dynamics
  of {V1} activity.
\newblock \emph{Proceedings of the National Academy of Sciences}, 117\penalty0
  (36):\penalty0 22494--22505, 2020.
\newblock \doi{10.1073/pnas.2005417117}.

\bibitem[Khalighi et~al.(2016)Khalighi, Sousa, Santos, and
  Nunes]{khalighi2016isruc}
Sirvan Khalighi, Teresa Sousa, Jos{\'e}~Moutinho Santos, and Urbano Nunes.
\newblock {ISRUC-Sleep}: A comprehensive public dataset for sleep researchers.
\newblock \emph{Computer Methods and Programs in Biomedicine}, 124:\penalty0
  180--192, 2016.
\newblock \doi{10.1016/j.cmpb.2015.10.013}.

\bibitem[Heeger(2017)]{heeger2017theory}
David~J Heeger.
\newblock Theory of cortical function.
\newblock \emph{Proceedings of the National Academy of Sciences}, 114\penalty0
  (8):\penalty0 1773--1782, 2017.
\newblock \doi{10.1073/pnas.1619788114}.

\bibitem[van~den Oord et~al.(2018)van~den Oord, Li, and Vinyals]{oord2018cpc}
A{\"a}ron van~den Oord, Yazhe Li, and Oriol Vinyals.
\newblock Representation learning with contrastive predictive coding.
\newblock \emph{arXiv preprint arXiv:1807.03748}, 2018.
\newblock URL \url{https://arxiv.org/abs/1807.03748}.

\bibitem[Yang et~al.(2023)Yang, Westover, and Sun]{yang2023biot}
Chaoqi Yang, M~Brandon Westover, and Jimeng Sun.
\newblock {BIOT}: Biosignal transformer for cross-data learning in the wild.
\newblock \emph{Advances in Neural Information Processing Systems},
  36:\penalty0 78240--78260, 2023.
\newblock URL
  \url{https://proceedings.neurips.cc/paper_files/paper/2023/hash/f6b30f3e2dd9cb53bbf2024402d02295-Abstract.html}.

\bibitem[Wang et~al.(2024)Wang, Yaari, Singh, Subramaniam, Rosenfarb, DeWitt,
  Misra, Madsen, Stone, Kreiman, Katz, Cases, and Barbu]{wang2024braintreebank}
Christopher Wang, Adam Yaari, Aaditya~K. Singh, Vighnesh Subramaniam, Dana
  Rosenfarb, Jan DeWitt, Pranav Misra, Joseph~R. Madsen, Scellig Stone, Gabriel
  Kreiman, Boris Katz, Ignacio Cases, and Andrei Barbu.
\newblock Brain treebank: Large-scale intracranial recordings from naturalistic
  language stimuli.
\newblock In \emph{Advances in Neural Information Processing Systems},
  volume~37, pages 96505--96540, 2024.
\newblock \doi{10.52202/079017-3060}.
\newblock Datasets and Benchmarks Track; arXiv:2411.08343.

\bibitem[Oganesian et~al.(2025)Oganesian, Hashemi, and
  Shanechi]{oganesian2025barista}
Lucine~L. Oganesian, Saba Hashemi, and Maryam~M. Shanechi.
\newblock {BaRISTA}: Brain scale informed spatiotemporal representation of
  human intracranial neural activity.
\newblock In \emph{Advances in Neural Information Processing Systems},
  volume~38, pages 185289--185322, 2025.
\newblock \doi{10.52202/085713-5573}.
\newblock arXiv:2512.12135.

\bibitem[Tangermann et~al.(2012)Tangermann, M{\"u}ller, Aertsen, Birbaumer,
  Braun, Brunner, Leeb, Mehring, Miller, M{\"u}ller-Putz, Nolte, Pfurtscheller,
  Preissl, Schalk, Schl{\"o}gl, Vidaurre, Waldert, and
  Blankertz]{tangermann2012bcic}
Michael Tangermann, Klaus-Robert M{\"u}ller, Ad~Aertsen, Niels Birbaumer,
  Christoph Braun, Clemens Brunner, Robert Leeb, Carsten Mehring, Kai~J.
  Miller, Gernot M{\"u}ller-Putz, Guido Nolte, Gert Pfurtscheller, Hubert
  Preissl, Gerwin Schalk, Alois Schl{\"o}gl, Carmen Vidaurre, Stephan Waldert,
  and Benjamin Blankertz.
\newblock Review of the {BCI} {C}ompetition {IV}.
\newblock \emph{Frontiers in Neuroscience}, 6:\penalty0 55, 2012.
\newblock \doi{10.3389/fnins.2012.00055}.

\bibitem[{Neural Engineering Data Consortium}()]{tuhcorpora}
{Neural Engineering Data Consortium}.
\newblock {Temple University EEG Corpus}: Downloads.
\newblock Temple University.
\newblock URL
  \url{https://isip.piconepress.com/projects/nedc/html/tuh_eeg/index.shtml}.
\newblock Accessed September 5, 2026.

\bibitem[Hong et~al.(2025)Hong, Mackellar, and Ghane]{hong2025samba}
Jiazhen Hong, Geoffrey Mackellar, and Soheila Ghane.
\newblock {SAMBA}: Toward a long-context {EEG} foundation model via spatial
  embedding and differential mamba.
\newblock arXiv preprint, 2025.
\newblock \href{https://arxiv.org/abs/2511.18571}{arXiv:2511.18571}.

\bibitem[Warden(2018)]{warden2018speech}
Pete Warden.
\newblock Speech commands: A dataset for limited-vocabulary speech recognition.
\newblock \emph{arXiv preprint arXiv:1804.03209}, 2018.
\newblock URL \url{https://arxiv.org/abs/1804.03209}.

\bibitem[Kidger et~al.(2020)Kidger, Morrill, Foster, and
  Lyons]{kidger2020neural}
Patrick Kidger, James Morrill, James Foster, and Terry Lyons.
\newblock Neural controlled differential equations for irregular time series.
\newblock \emph{Advances in neural information processing systems},
  33:\penalty0 6696--6707, 2020.
\newblock URL
  \url{https://proceedings.neurips.cc/paper/2020/hash/4a5876b450b45371f6cfe5047ac8cd45-Abstract.html}.

\bibitem[Romero et~al.(2022{\natexlab{a}})Romero, Kuzina, Bekkers, Tomczak, and
  Hoogendoorn]{romero2022ckconv}
David~W Romero, Anna Kuzina, Erik~J Bekkers, Jakub~M Tomczak, and Mark
  Hoogendoorn.
\newblock {CKConv}: Continuous kernel convolution for sequential data.
\newblock In \emph{International Conference on Learning Representations},
  2022{\natexlab{a}}.
\newblock URL \url{https://arxiv.org/abs/2102.02611}.

\bibitem[Romero et~al.(2022{\natexlab{b}})Romero, Bruintjes, Bekkers, Tomczak,
  Hoogendoorn, and van Gemert]{romero2022flexconv}
David~W Romero, Robert-Jan Bruintjes, Erik~J Bekkers, Jakub~M Tomczak, Mark
  Hoogendoorn, and Jan~C van Gemert.
\newblock {FlexConv}: Continuous kernel convolutions with differentiable kernel
  sizes.
\newblock In \emph{International Conference on Learning Representations},
  2022{\natexlab{b}}.
\newblock URL \url{https://arxiv.org/abs/2110.08059}.

\bibitem[El~Ouahidi et~al.(2026)]{reve2025code}
Yassine El~Ouahidi et~al.
\newblock {REVE} official implementation: {BCI-IV-2a} preprocessing.
\newblock GitHub repository, commit 06a7059, 2026.
\newblock URL
  \url{https://github.com/elouayas/reve_eeg/blob/06a7059a07c3dabd80aee60c3dbc1eca4bdbe1c7/preprocessing/preprocessing_bciciv2a.py}.

\bibitem[He and Wu(2020)]{he2020euclidean}
He~He and Dongrui Wu.
\newblock Transfer learning for brain--computer interfaces: A {E}uclidean space
  data alignment approach.
\newblock \emph{IEEE Transactions on Biomedical Engineering}, 67\penalty0
  (2):\penalty0 399--410, 2020.
\newblock \doi{10.1109/TBME.2019.2913914}.

\bibitem[Baevski et~al.(2020)Baevski, Zhou, Mohamed, and
  Auli]{baevski2020wav2vec}
Alexei Baevski, Henry Zhou, Abdelrahman Mohamed, and Michael Auli.
\newblock wav2vec 2.0: A framework for self-supervised learning of speech
  representations.
\newblock \emph{Advances in neural information processing systems},
  33:\penalty0 12449--12460, 2020.
\newblock URL
  \url{https://proceedings.neurips.cc/paper/2020/hash/92d1e1eb1cd6f9fba3227870bb6d7f07-Abstract.html}.

\bibitem[Hsu et~al.(2021)Hsu, Bolte, Tsai, Lakhotia, Salakhutdinov, and
  Mohamed]{hsu2021hubert}
Wei-Ning Hsu, Benjamin Bolte, Yao-Hung~Hubert Tsai, Kushal Lakhotia, Ruslan
  Salakhutdinov, and Abdelrahman Mohamed.
\newblock {HuBERT}: Self-supervised speech representation learning by masked
  prediction of hidden units.
\newblock \emph{IEEE/ACM Transactions on Audio, Speech, and Language
  Processing}, 29:\penalty0 3451--3460, 2021.
\newblock \doi{10.1109/TASLP.2021.3122291}.

\bibitem[Assran et~al.(2023)Assran, Duval, Misra, Bojanowski, Vincent, Rabbat,
  LeCun, and Ballas]{assran2023ijepa}
Mahmoud Assran, Quentin Duval, Ishan Misra, Piotr Bojanowski, Pascal Vincent,
  Michael Rabbat, Yann LeCun, and Nicolas Ballas.
\newblock Self-supervised learning from images with a joint-embedding
  predictive architecture.
\newblock In \emph{Proceedings of the IEEE/CVF Conference on Computer Vision
  and Pattern Recognition}, pages 15619--15629, 2023.
\newblock URL
  \url{https://openaccess.thecvf.com/content/CVPR2023/html/Assran_Self-Supervised_Learning_From_Images_With_a_Joint-Embedding_Predictive_Architecture_CVPR_2023_paper.html}.

\bibitem[Bardes et~al.(2024)Bardes, Garrido, Ponce, Chen, Rabbat, LeCun,
  Assran, and Ballas]{bardes2024vjepa}
Adrien Bardes, Quentin Garrido, Jean Ponce, Xinlei Chen, Michael Rabbat, Yann
  LeCun, Mahmoud Assran, and Nicolas Ballas.
\newblock Revisiting feature prediction for learning visual representations
  from video.
\newblock \emph{arXiv preprint arXiv:2404.08471}, 2024.
\newblock URL \url{https://arxiv.org/abs/2404.08471}.

\bibitem[Yadav and Tan(2024)]{yadav2024audiomamba}
Sarthak Yadav and Zheng-Hua Tan.
\newblock Audio mamba: Selective state spaces for self-supervised audio
  representations.
\newblock In \emph{Interspeech 2024}, pages 552--556, 2024.
\newblock \doi{10.21437/Interspeech.2024-1274}.
\newblock arXiv:2406.02178.

\bibitem[Gong et~al.(2022)Gong, Lai, Chung, and Glass]{gong2022ssast}
Yuan Gong, Cheng-I~Jeff Lai, Yu-An Chung, and James Glass.
\newblock {SSAST}: Self-supervised audio spectrogram transformer.
\newblock In \emph{Proceedings of the AAAI Conference on Artificial
  Intelligence}, volume~36, pages 10699--10709, 2022.
\newblock \doi{10.1609/aaai.v36i10.21315}.

\bibitem[Mehari and Strodthoff(2023)]{mehari2023quantitative}
Temesgen Mehari and Nils Strodthoff.
\newblock Towards quantitative precision for {ECG} analysis: Leveraging state
  space models, self-supervision and patient metadata.
\newblock \emph{IEEE Journal of Biomedical and Health Informatics}, 27\penalty0
  (11):\penalty0 5326--5334, 2023.
\newblock \doi{10.1109/JBHI.2023.3310989}.

\bibitem[Al-Masud and Strodthoff(2026)]{almasud2026pretraining}
M.~A. Al-Masud and Nils Strodthoff.
\newblock Pretraining strategies and scaling for {ECG} foundation models: A
  systematic study.
\newblock arXiv preprint, 2026.
\newblock \href{https://arxiv.org/abs/2605.12241}{arXiv:2605.12241}.

\bibitem[Wu et~al.(2026)Wu, Deep, Lehman, Kapoor, Zhao, Krishnan, Clifford, and
  Lehman]{wu2026sls4wave}
Feng Wu, Harsh Deep, Eric Lehman, Sanyam Kapoor, Guoshuai Zhao, Rahul~G.
  Krishnan, Gari Clifford, and Li-wei~H. Lehman.
\newblock {SL-S4Wave}: Self-supervised learning of physiological waveforms with
  structured state space models.
\newblock \emph{Transactions on Machine Learning Research}, 2026.
\newblock URL \url{https://openreview.net/forum?id=km0xS3jZeO}.
\newblock \href{https://arxiv.org/abs/2606.19888}{arXiv:2606.19888}.

\bibitem[Broustail et~al.(2026)Broustail, Tegon, Ingolfsson, Li, and
  Benini]{broustail2026lumamba}
Dana{\'e} Broustail, Anna Tegon, Thorir~Mar Ingolfsson, Yawei Li, and Luca
  Benini.
\newblock {LuMamba}: Latent unified mamba for electrode topology-invariant and
  efficient {EEG} modeling.
\newblock arXiv preprint, 2026.
\newblock \href{https://arxiv.org/abs/2603.19100}{arXiv:2603.19100}.

\bibitem[Puah et~al.(2025)Puah, Goh, Zhang, Ye, Chan, Lim, Fong, Woon, and
  Guan]{puah2025eegdm}
Jia~Hong Puah, Sim~Kuan Goh, Ziwei Zhang, Zixuan Ye, Chow~Khuen Chan,
  Kheng~Seang Lim, Si~Lei Fong, Kok~Sin Woon, and Cuntai Guan.
\newblock {EEGDM}: {EEG} representation learning via generative diffusion
  model.
\newblock arXiv preprint, 2025.
\newblock \href{https://arxiv.org/abs/2508.14086}{arXiv:2508.14086}.

\bibitem[Durgam et~al.(2026)Durgam, Siddiqui, Chan-Santiago, Fu, Gireesh, and
  Shah]{durgam2026cambrain}
Abhilash Durgam, Nyle Siddiqui, Jeffrey~A. Chan-Santiago, Qiushi Fu, Elakkat~D.
  Gireesh, and Mubarak Shah.
\newblock {CaMBrain}: Real-time, continuous {EEG} inference with causal state
  space models.
\newblock arXiv preprint, 2026.
\newblock \href{https://arxiv.org/abs/2605.28792}{arXiv:2605.28792}.

\bibitem[Zhang et~al.(2023)Zhang, Yuan, Yang, Chen, Wang, and
  Li]{zhang2023brant}
Daoze Zhang, Zhizhang Yuan, Yang Yang, Junru Chen, Jingjing Wang, and Yafeng
  Li.
\newblock Brant: Foundation model for intracranial neural signal.
\newblock In \emph{Advances in Neural Information Processing Systems},
  volume~36, pages 26304--26321, 2023.
\newblock \doi{10.52202/075280-1144}.

\bibitem[Wang et~al.(2023)Wang, Subramaniam, Yaari, Kreiman, Katz, Cases, and
  Barbu]{wang2023brainbert}
Christopher Wang, Vighnesh Subramaniam, Adam~Uri Yaari, Gabriel Kreiman, Boris
  Katz, Ignacio Cases, and Andrei Barbu.
\newblock {BrainBERT}: Self-supervised representation learning for intracranial
  recordings.
\newblock In \emph{International Conference on Learning Representations}, 2023.
\newblock \href{https://arxiv.org/abs/2302.14367}{arXiv:2302.14367}.

\bibitem[Chau et~al.(2025)Chau, Wang, Talukder, Subramaniam, Soedarmadji, Yue,
  Katz, and Barbu]{chau2025popt}
Geeling Chau, Christopher Wang, Sabera Talukder, Vighnesh Subramaniam,
  Saraswati Soedarmadji, Yisong Yue, Boris Katz, and Andrei Barbu.
\newblock Population transformer: Learning population-level representations of
  neural activity.
\newblock In \emph{International Conference on Learning Representations}, 2025.
\newblock \href{https://arxiv.org/abs/2406.03044}{arXiv:2406.03044}.

\end{thebibliography}

\clearpage
\appendix

\section{Complete results by benchmark}
\label{app:additional-results}

This appendix reports the complete results behind the main-text comparisons.
To measure the effect of SSL, we train a second model from the same
starting weights without pretraining; tables label this control
\textit{paired init.} Published values use the evaluation procedures of
their source papers. Relevant differences from our procedures are
listed with each dataset in Appendix~\ref{app:datasets}.

\paragraph{Metrics.}
Binary tasks report AUROC, the probability that a positive example is
scored higher than a negative one, and balanced accuracy (B-ACC), the mean
recall across classes. ISRUC and BCI-IV-2a use B-ACC as their primary
metric; Speech Commands (SC10) reports accuracy.

\subsection{TUAB}

TUAB contains clinical scalp EEG recordings from Temple University
Hospital, labeled as normal or abnormal~\cite{tuhcorpora}. Our preparation contains
2{,}076 patients for training and validation and 253 for testing.
We feed the EEG signals to the RQF encoder and compare the windowed and
dense prediction objectives of Section~\ref{sec:method}, then train a
readout to distinguish normal from abnormal recordings.

\begin{figure}[!htbp]
  \centering
  \includegraphics[width=\textwidth]{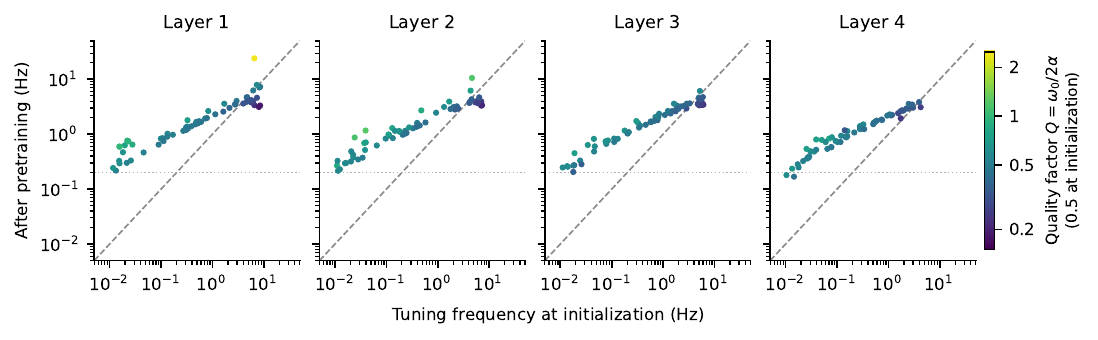}
  \caption{\textbf{Pretraining retunes the TUAB filters.} Tuning frequency
  $\omega_0/2\pi$ of each of the 64 filters per layer after windowed
  pretraining, against its value at initialization; the dashed line marks
  no change and the dotted line 0.2\,Hz. Color gives the quality factor
  $Q=\omega_0/2\alpha$, which is 0.5 for every filter at initialization,
  the band-pass critical value~\cite{rqf2026}. Pretraining moves 246 of
  the 256 filters into 0.2 to 5\,Hz; 98 start below 0.2\,Hz and 2 remain
  there. In each layer, 31 to 50\% of the filters end with $Q$ above 0.5.}
  \label{fig:filters}
\end{figure}

\paragraph{Full-data comparison.}
The windowed objective leads dense prediction more clearly in AUROC than
in balanced accuracy. In a separate comparison on 100-s segments with both prediction
horizons fixed at two seconds, the encoders frozen, and the default
fine-tuning settings of Appendix~\ref{app:training-protocols}, the two objectives achieve $0.8978\pm0.0089$ and
$0.8888\pm0.0027$ AUROC, respectively; their balanced accuracies differ
by only 0.19 points, within the variation across seeds. The windowed
results in Tables~\ref{tab:tuab} and~\ref{tab:tuab100} use a one-second
prediction horizon and the TUAB and ISRUC fine-tuning settings of
Appendix~\ref{app:training-protocols}.
The test set matches the official evaluation partition in size and class
balance. The available metadata do not allow us to verify this
correspondence patient by patient.
Figure~\ref{fig:filters} shows how windowed pretraining retunes the
filters of the TUAB encoder.

\begin{table}[!htbp]
  \centering
  \caption{TUAB per-segment test metrics on 10-s segments. Published
  values are from CBraMod, REVE, and
  SAMBA~\cite{wang2025cbramod,reve2025,hong2025samba}; they use 200-Hz
  inputs except SAMBA, which uses 128\,Hz as we do. Our results use three
  downstream seeds from a fixed pretraining checkpoint, and paired init.\
  starts from the epoch-zero weights of that checkpoint without
  pretraining. Parameter counts follow the real-scalar convention of
  Table~\ref{tab:params}. Bold marks the configuration plotted in
  Figure~\ref{fig:pareto}; with full fine-tuning, paired init.\ is level
  with it.}
  \label{tab:tuab}
  \small
  \setlength{\tabcolsep}{1.8pt}
  \begin{tabular}{lrlcrr}
    \toprule
    Model & Params & Context & $n$ & AUROC & B-ACC \\
    \midrule
    EEGNet & 3k & 10\,s/200\,Hz & 5 & 0.8412 $\pm$ .0031 & 0.7642 $\pm$ .0036 \\
    EEGConformer & 550k & 10\,s/200\,Hz & 5 & 0.8445 $\pm$ .0038 & 0.7758 $\pm$ .0049 \\
    SPaRCNet & 790k & 10\,s/200\,Hz & 5 & 0.8676 $\pm$ .0012 & 0.7896 $\pm$ .0018 \\
    ContraWR & 1.6M & 10\,s/200\,Hz & 5 & 0.8456 $\pm$ .0074 & 0.7746 $\pm$ .0041 \\
    CNN-Transformer & 3.2M & 10\,s/200\,Hz & 5 & 0.8461 $\pm$ .0013 & 0.7777 $\pm$ .0022 \\
    FFCL & 2.4M & 10\,s/200\,Hz & 5 & 0.8569 $\pm$ .0051 & 0.7848 $\pm$ .0038 \\
    ST-Transformer & 3.5M & 10\,s/200\,Hz & 5 & 0.8707 $\pm$ .0019 & 0.7966 $\pm$ .0023 \\
    \midrule
    BIOT & 3.2M & 10\,s/200\,Hz & 5 & 0.8815 $\pm$ .0043 & 0.7959 $\pm$ .0057 \\
    LaBraM-Base & 5.8M & 10\,s/200\,Hz & 5 & 0.9022 $\pm$ .0009 & 0.8140 $\pm$ .0019 \\
    LaBraM-Large & 46M & 10\,s/200\,Hz & 5 & 0.9127 $\pm$ .0005 & 0.8226 $\pm$ .0015 \\
    LaBraM-Huge & 369M & 10\,s/200\,Hz & 5 & 0.9162 $\pm$ .0016 & 0.8258 $\pm$ .0011 \\
    CBraMod & 4.0M & 10\,s/200\,Hz & 5 & 0.9227 $\pm$ .0011 & 0.8289 $\pm$ .0022 \\
    REVE-Base & 69M & 10\,s/200\,Hz & n.r. & 0.9245 $\pm$ .0013 & 0.8315 $\pm$ .0014 \\
    SAMBA-10s & 1.0M & 10\,s/128\,Hz & n.r. & 0.8887 $\pm$ .0004 & 0.8150 $\pm$ .0037 \\
    \midrule
    Ours --- paired init., frozen & 110k & 10\,s/128\,Hz & 3 & 0.8643 $\pm$ .0009 & 0.7769 $\pm$ .0023 \\
    Ours --- windowed SSL, frozen & 110k & 10\,s/128\,Hz & 3 & 0.8939 $\pm$ .0010 & 0.8198 $\pm$ .0015 \\
    Ours --- paired init., full-FT & 110k & 10\,s/128\,Hz & 3 & 0.9046 $\pm$ .0017 & 0.8268 $\pm$ .0021 \\
    Ours --- windowed SSL, full-FT & 110k & 10\,s/128\,Hz & 3 & \textbf{0.9040 $\pm$ .0019} & \textbf{0.8273 $\pm$ .0029} \\
    \bottomrule
  \end{tabular}
\end{table}

\paragraph{Fine-tuned performance depends most on a few alpha- and beta-tuned filters.}
On the TUAB validation set, we set to zero the states of all filters whose
tuning frequency lies in one EEG band, in every layer of a trained model,
and compare the AUROC drop with that of removing as many randomly chosen
filters from the same layers (Table~\ref{tab:band-ablation}). In the fully
fine-tuned model, silencing its 5 to 7 alpha-tuned filters lowers AUROC by
13.4 points, against 0.5 for random filters, and its single beta-tuned
filter costs 7.7 points. Filters tuned below 1\,Hz matter less than random
ones. Fine-tuning moves filters into the theta and alpha bands (28 to 33
and 5 to 7, against 16 and 1 in the frozen encoder). The models are not
retrained after silencing, so the drops measure how much each model relies
on these filters, not whether they are necessary.

\begin{table}[!htbp]
  \centering
  \caption{TUAB validation AUROC drop (points) when the states of all
  filters tuned to one band are set to zero in every layer, mean over three
  downstream seeds. Excess is the drop minus that of removing as many
  random filters from the same layers (20 draws); brackets are 95\%
  intervals over patients (5{,}000 draws). No filter is tuned above
  30\,Hz.}
  \label{tab:band-ablation}
  \small
  \setlength{\tabcolsep}{3pt}
  \begin{tabular}{lrrrrrr}
    \toprule
    & \multicolumn{3}{c}{Windowed SSL, frozen} & \multicolumn{3}{c}{Windowed SSL, full-FT} \\
    \cmidrule(lr){2-4}\cmidrule(lr){5-7}
    Band & Filters & Drop & Excess & Filters & Drop & Excess \\
    \midrule
    Below 1\,Hz & 90 & 9.6 & $-7.7$ [$-9.2$, $-6.1$] & 87--89 & 4.2 & $-5.7$ [$-7.0$, $-4.5$] \\
    Delta (1--4\,Hz) & 148 & 32.2 & 9.6 [5.2, 14.1] & 130--132 & 11.3 & $-5.7$ [$-7.2$, $-4.3$] \\
    Theta (4--8\,Hz) & 16 & 6.7 & 2.8 [1.5, 4.2] & 28--33 & 7.1 & 2.2 [0.2, 4.2] \\
    Alpha (8--13\,Hz) & 1 & 2.7 & 2.5 [1.2, 3.8] & 5--7 & 13.4 & 12.9 [10.2, 15.8] \\
    Beta (13--30\,Hz) & 1 & 3.1 & 3.0 [1.7, 4.4] & 1 & 7.7 & 7.4 [5.7, 9.4] \\
    \bottomrule
  \end{tabular}
\end{table}

\paragraph{100-s segments.}
With 100-s segments, the context of SAMBA-100s, windowed pretraining
followed by full fine-tuning achieves $0.9247 \pm 0.0022$ test AUROC
(Table~\ref{tab:tuab100}). Its paired initialization achieves
$0.9174 \pm 0.0012$, so pretraining adds 0.7 points (95\% interval over
test patients, 0.1 to 1.4). The same settings train on the whole segment
without random cropping.

\begin{table}[!htbp]
  \centering
  \caption{TUAB per-segment test metrics on 100-s segments at 128\,Hz,
  the configuration of SAMBA-100s~\cite{hong2025samba}. Our results use
  three downstream seeds from the pretraining checkpoint of
  Table~\ref{tab:tuab}; both are fully fine-tuned, and paired init.\
  starts from the same epoch-zero weights without pretraining. Bold marks
  our strongest configuration.}
  \label{tab:tuab100}
  \small
  \setlength{\tabcolsep}{1.8pt}
  \begin{tabular}{lrlcrr}
    \toprule
    Model & Params & Context & $n$ & AUROC & B-ACC \\
    \midrule
    SAMBA-100s & 1.0M & 100\,s/128\,Hz & n.r. & 0.9054 $\pm$ .0044 & 0.8264 $\pm$ .0017 \\
    \midrule
    Ours --- paired init., full-FT & 110k & 100\,s/128\,Hz & 3 & 0.9174 $\pm$ .0012 & 0.8357 $\pm$ .0009 \\
    Ours --- windowed SSL, full-FT & 110k & 100\,s/128\,Hz & 3 & \textbf{0.9247 $\pm$ .0022} & \textbf{0.8426 $\pm$ .0025} \\
    \bottomrule
  \end{tabular}
\end{table}

\paragraph{TUAB label efficiency.}
Pretraining helps most when only 1\% of the training labels are
available. With full fine-tuning, the benefit shrinks as labels increase
and disappears with all labels. Frozen pretrained encoders score 2.3 to
3.4 points above their paired-initialization controls at every label
budget. Each comparison starts from the same weights and uses the same
validation and test patients. Intervals over test patients exclude zero
for full fine-tuning at 1\% and for frozen encoders at 25\% and 100\%.
At 1\%, one of three seeds is worse with pretraining in both settings.

\begin{table}[!htbp]
  \centering
  \caption{TUAB label efficiency for windowed SSL and its paired
  initialization (test AUROC on 10-s segments; mean$\pm$std over three
  downstream seeds from a fixed checkpoint). All results use the
  110k-parameter model of Table~\ref{tab:params}. Differences are
  computed from unrounded values. Bold marks the strongest configuration
  at each budget.}
  \label{tab:tuab_label_efficiency}
  \small
  \setlength{\tabcolsep}{2.0pt}
  \begin{tabular}{lccccc}
    \toprule
    Configuration & 1\% & 5\% & 10\% & 25\% & 100\% \\
    \midrule
    Full-FT, windowed SSL & \textbf{0.7001 $\pm$ .0722} & \textbf{0.8041 $\pm$ .0030} & \textbf{0.8395 $\pm$ .0054} & \textbf{0.8782 $\pm$ .0014} & 0.9040 $\pm$ .0019 \\
    Full-FT, paired init. & 0.6509 $\pm$ .0230 & 0.7822 $\pm$ .0035 & 0.8271 $\pm$ .0048 & 0.8736 $\pm$ .0020 & \textbf{0.9046 $\pm$ .0017} \\
    Full-FT difference (pp) & +4.91 & +2.19 & +1.24 & +0.45 & $-$0.06 \\
    \midrule
    Frozen, windowed SSL & 0.6845 $\pm$ .0718 & 0.7906 $\pm$ .0069 & 0.8245 $\pm$ .0038 & 0.8603 $\pm$ .0013 & 0.8939 $\pm$ .0010 \\
    Frozen, paired init. & 0.6501 $\pm$ .0060 & 0.7624 $\pm$ .0096 & 0.8014 $\pm$ .0017 & 0.8353 $\pm$ .0025 & 0.8643 $\pm$ .0009 \\
    Frozen difference (pp) & +3.44 & +2.82 & +2.32 & +2.50 & +2.96 \\
    \bottomrule
  \end{tabular}
\end{table}

\subsection{ISRUC-SG1}

ISRUC-Sleep contains overnight recordings collected at the Sleep Medicine
Centre of Coimbra University Hospital; subgroup~I has one recording from
each of 100 adults undergoing sleep assessment~\cite{khalighi2016isruc}.
We use six EEG channels from 99 participants, excluding one recording
that lacks two required channels, with 79/10/10 participants for training,
validation, and testing. We pretrain the RQF encoder with windowed future
prediction (Section~\ref{sec:method}) and fine-tune it to assign one of
five sleep stages to each 30-second epoch.

\paragraph{Epoch context.}
Adding neighboring epochs improves balanced accuracy from
$0.7746\pm0.0134$ to $0.7990\pm0.0087$ when both models use the
default fine-tuning settings of Appendix~\ref{app:training-protocols}.
With the TUAB and ISRUC fine-tuning settings of the same appendix, the contextual model achieves
$0.8223\pm0.0028$ (Table~\ref{tab:isruc}).
Both RQF models
first encode each 30-second epoch independently. The epoch-local model
predicts from that representation alone; the contextual model passes the
20 representations through separate forward and reverse one-layer RQFs
before predicting all 20 labels. The published baselines also use 20-epoch
context, but CBraMod's evaluation combines the epoch representations with
a shared one-layer Transformer~\cite{wang2025cbramod}. Our contextual model
uses RQFs for both epoch encoding and context, so both components differ
from those baselines.

\begin{table}[!htbp]
  \centering
  \caption{ISRUC-SG1 five-class sleep-staging test balanced accuracy.
  Published baselines except REVE-Base come from CBraMod Table~7 and use
  its shared Transformer over 20 epochs~\cite{wang2025cbramod}. Their
  parameter counts omit the Transformer, whereas ours include the complete
  downstream model. REVE reports 20-epoch
  sequence prediction without specifying its context module, so the table
  marks its context as reported~\cite{reve2025}. Our split excludes subject 8,
  whose F3 and F4 channels are missing, so we train on 79 subjects instead
  of 80; validation and test subjects match CBraMod's. Our results are
  mean$\pm$std over three downstream seeds from a fixed pretraining
  checkpoint, and paired init.\ starts from the epoch-zero weights of that
  checkpoint without pretraining. Bold marks the configuration plotted in
  Figure~\ref{fig:pareto}, not the best value in the column.}
  \label{tab:isruc}
  \small
  \setlength{\tabcolsep}{3.0pt}
  \begin{tabular}{lrlr}
    \toprule
    Model & Params & Epoch context & B-ACC \\
    \midrule
    EEGNet & 3k & 20 $\times$ 30\,s & 0.7154 $\pm$ .0121 \\
    EEGConformer & 550k & 20 $\times$ 30\,s & 0.7400 $\pm$ .0133 \\
    SPaRCNet & 790k & 20 $\times$ 30\,s & 0.7487 $\pm$ .0075 \\
    ContraWR & 1.6M & 20 $\times$ 30\,s & 0.7402 $\pm$ .0126 \\
    CNN-Transformer & 3.2M & 20 $\times$ 30\,s & 0.7363 $\pm$ .0087 \\
    FFCL & 2.4M & 20 $\times$ 30\,s & 0.7277 $\pm$ .0182 \\
    ST-Transformer & 3.5M & 20 $\times$ 30\,s & 0.7381 $\pm$ .0205 \\
    DeepSleepNet & 21M & 20 $\times$ 30\,s & 0.7419 $\pm$ .0144 \\
    U-Sleep & 1.1M & 20 $\times$ 30\,s & 0.7586 $\pm$ .0116 \\
    BIOT & 3.2M & 20 $\times$ 30\,s & 0.7527 $\pm$ .0121 \\
    LaBraM-Base & 5.8M & 20 $\times$ 30\,s & 0.7633 $\pm$ .0102 \\
    CBraMod & 4.0M & 20 $\times$ 30\,s & 0.7865 $\pm$ .0110 \\
    REVE-Base & 69M & reported: 20 $\times$ 30\,s & 0.7819 $\pm$ .0078 \\
    \midrule
    Ours --- paired init., full-FT & 224k & 20 $\times$ 30\,s &
      0.8252 $\pm$ .0016 \\
    \textbf{Ours --- windowed SSL, full-FT} & 224k & 20 $\times$ 30\,s &
      \textbf{0.8223 $\pm$ .0028} \\
    \bottomrule
  \end{tabular}
\end{table}

\FloatBarrier
\subsection{Speech Commands}

Speech Commands is a collection of short spoken words recorded by
volunteers, released by Google for keyword-recognition
research~\cite{warden2018speech}. We use the ten-command subset of
34{,}975 exactly one-second clips, with a class-stratified 70/15/15
training, validation, and test split. The RQF encoder receives the audio
waveform and learns through the same windowed or dense prediction
objective as in Section~\ref{sec:method}; the downstream task identifies
the spoken command.

\paragraph{Dense and windowed prediction.}
On these non-neural signals, both
objectives approach the strongest published compact models after full
fine-tuning. With the encoder frozen, windowed prediction retains a
1.0-point advantage over dense prediction, while both remain far above
their paired-initialization control (Table~\ref{tab:sc10}).

\begin{table}[!htbp]
  \centering
  \caption{SC10 ten-class test accuracy on the exact 34{,}975 one-second
  clips. Our models use raw waveforms and the same per-timestep readout.
  Our values are mean$\pm$std over three downstream seeds; bold marks our
  strongest configuration, not the best value in the column.}
  \label{tab:sc10}
  \small
  \setlength{\tabcolsep}{3.0pt}
  \begin{tabular}{lrlcr}
    \toprule
    Model & Params & Input & $n$ & Accuracy \\
    \midrule
    Neural CDE~\cite{kidger2020neural} & 89k & MFCC & 5 & 0.898 $\pm$ .025 \\
    CKConv~\cite{romero2022ckconv} & 100k & MFCC & n.r. & 0.9527 \\
    FlexConv~\cite{romero2022flexconv} & 239k & MFCC & n.r. & 0.9773 \\
    \midrule
    Ours --- paired init., frozen & 69.3k & raw & 3 & 0.6731 $\pm$ .0043 \\
    Ours --- paired init., full-FT & 69.3k & raw & 3 & 0.9666 $\pm$ .0034 \\
    Ours --- dense SSL, frozen & 69.3k & raw & 3 & 0.9408 $\pm$ .0009 \\
    Ours --- windowed SSL, frozen & 69.3k & raw & 3 & 0.9503 $\pm$ .0012 \\
    Ours --- dense SSL, full-FT & 69.3k & raw & 3 & \textbf{0.9723 $\pm$ .0029} \\
    Ours --- windowed SSL, full-FT & 69.3k & raw & 3 & 0.9717 $\pm$ .0010 \\
    \bottomrule
  \end{tabular}
\end{table}

\paragraph{Phase-inclusive target diagnostic.}
In a separate SC10 diagnostic with an EMA target encoder, we compared
power prediction alone with an additional phase-increment prediction
head. Both frozen encoders were evaluated with the same diagnostic
probe. The phase head achieved 0.885 validation retrieval accuracy, but
adding it reduced downstream validation macro recall from 0.8086 to
0.7045 with 5\% of the labels and from 0.9040 to 0.8800 with all labels.
This comparison covers one seed and does not test other ways of
representing phase.

\subsection{BrainTreebank}

BrainTreebank records intracranial EEG from ten patients undergoing
epilepsy monitoring at Boston Children's Hospital while watching
movies~\cite{wang2024braintreebank}. We use 17 sessions for SSL and seven
other sessions for downstream evaluation, with seven participants shared
between these sets. We apply the windowed and dense objectives of
Section~\ref{sec:method} to the intracranial signals and train the
downstream readout to detect sentence onsets. With the encoder frozen,
windowed pretraining achieved $0.8822\pm0.0011$ AUROC, compared with
$0.8247\pm0.0064$ for paired initialization (Table~\ref{tab:btb}).

\paragraph{Evaluation protocol and published reference points.}
We use BaRISTA's chronological fold generator, which places validation
blocks between training and test blocks, with five folds in each of its
seven held-out sessions~\cite{oganesian2025barista}. Published reference
values come from BaRISTA's chronological evaluation (its Appendix Table~16),
which uses the same sessions and folds. The
evaluation sessions are held out from SSL, but each of the seven
evaluation patients also contributes other sessions to pretraining.

\paragraph{Controlled comparisons.}
The controlled matrix separates the pretext objective from the way the
encoder is used downstream. Table~\ref{tab:btb} uses the same architecture
and seven sessions with five folds each while varying pretraining and
whether the encoder is frozen. Table~\ref{tab:btb_mse} tests the same transfer modes
after regression pretraining; raw-voltage and STFT probes calibrate the
task without a learned encoder. We retrain the controlled baselines on
our folds.

Both non-neural baselines use L2-regularized logistic regression. The
inputs are either standardized voltage samples or STFT magnitudes
computed with a 512-sample Hann window, 384-sample overlap, and
frequencies up to 150\,Hz; regularization is selected on validation data.

\begin{table}[!htbp]
  \centering
  \caption{BrainTreebank sentence-onset test AUROC over seven held-out
  sessions with five chronological folds each. For each seed, we average
  the 35 session-by-fold results, then report mean$\pm$std across
  five downstream seeds. Published values are
  mean$\pm$s.e.m. from BaRISTA's chronological evaluation. Bold marks our strongest configuration, not
  the best value in the column.}
  \label{tab:btb}
  \small
  \begin{tabular}{lrcc}
    \toprule
    Model & Params & Seeds & AUROC \\
    \midrule
    Brant & 500M & 5 & 0.772 $\pm$ .009 \\
    PopT{+}BrainBERT & 20M & 5 & 0.776 $\pm$ .009 \\
    BaRISTA (channels) & 1.0M & 5 & 0.778 $\pm$ .009 \\
    BaRISTA (parcels) & 1.0M & 5 & 0.853 $\pm$ .007 \\
    \midrule
    Ours --- paired init., frozen & 293k & 5 & 0.8247 $\pm$ .0064 \\
    Ours --- paired init., full-FT & 293k & 5 & 0.8459 $\pm$ .0077 \\
    Ours --- dense InfoNCE SSL, frozen & 293k & 5 & 0.8670 $\pm$ .0055 \\
    Ours --- dense InfoNCE SSL, full-FT & 293k & 5 & 0.8529 $\pm$ .0042 \\
    Ours --- windowed InfoNCE SSL, frozen & 293k & 5 & \textbf{0.8822 $\pm$ .0011} \\
    Ours --- windowed InfoNCE SSL, full-FT & 293k & 5 & 0.8706 $\pm$ .0022 \\
    \bottomrule
  \end{tabular}
\end{table}

\begin{table}[!htbp]
  \centering
  \caption{Controlled BrainTreebank baselines and windowed-MSE diagnostic
  across the same seven sessions with five folds each (test AUROC). Logistic
  results are deterministic fits, mean$\pm$std across these 35 results; MSE results
  are mean$\pm$std over five downstream seeds.}
  \label{tab:btb_mse}
  \small
  \begin{tabular}{lr}
    \toprule
    Configuration & Test AUROC \\
    \midrule
    Logistic regression, raw voltage & 0.7633 $\pm$ .1091 \\
    Logistic regression, STFT magnitude & 0.8621 $\pm$ .0807 \\
    \midrule
    Windowed MSE SSL, frozen & 0.8728 $\pm$ .0043 \\
    Windowed MSE SSL, full-FT & 0.8727 $\pm$ .0030 \\
    \bottomrule
  \end{tabular}
\end{table}

Six contrasts were fixed before the test set was evaluated, each paired over the
35 session-by-fold results after averaging the five downstream seeds.
Their 95\% intervals resample the seven sessions with replacement, keeping
each session's folds together (5{,}000 draws). Frozen windowed SSL exceeds
its fully fine-tuned version by 1.2 points (interval 0.2 to 2.3), frozen
dense SSL by 1.5 points (0.6 to 2.6), the STFT probe by 2.0 points (0.04
to 4.2), and frozen paired initialization by 5.7 points (2.8 to 9.4).
Paired initialization orders the other way, with full fine-tuning 2.1
points above its frozen version (0.4 to 4.1). After MSE pretraining, the
frozen and fully fine-tuned encoders are level (0.0 points, $-0.7$ to
$0.7$).

\subsection{BCI-IV-2a}

BCI-IV-2a contains scalp EEG recorded at Graz University of Technology
from nine participants imagining movements of the left hand, right hand,
feet, or tongue~\cite{tangermann2012bcic}. We use all 5{,}184 trials, with
five participants for training and two each for validation and testing,
following REVE's released split~\cite{reve2025code}. We apply windowed
future prediction (Section~\ref{sec:method}) either to the training
participants' EEG or to a 3{,}768-hour subset of TUEG, the larger clinical
EEG archive from Temple University Hospital~\cite{tuhcorpora}, before fine-tuning for
motor-imagery classification.

\paragraph{Cross-corpus comparison.}
We use Euclidean alignment to reduce differences between recording
sessions, estimating a whitening matrix for each session without using
its labels~\cite{he2020euclidean}. With this preprocessing, our RQF trained
directly on labeled data performs within 0.9 balanced-accuracy points of
the models pretrained on BCI-IV-2a or TUEG. Omitting alignment lowers the
accuracy of both pretrained models.
Published values come from REVE's
benchmark. REVE applies Euclidean alignment in its own
preprocessing~\cite{reve2025}, so we compare its result with our aligned results.

\begin{table}[!htbp]
  \centering
  \caption{BCI-IV-2a four-class motor-imagery test balanced accuracy.
  Published values come from REVE Table~12~\cite{reve2025}, which does not
  report repeat counts. REVE applies Euclidean alignment (EA) in its own
  preprocessing and does not state the preprocessing behind its baselines. Our models use full
  fine-tuning and report mean$\pm$std over three seeds; bold marks our
  strongest configuration, not the best value in the column.}
  \label{tab:bcic}
  \small
  \setlength{\tabcolsep}{2.4pt}
  \begin{tabular}{lrlcr}
    \toprule
    Model & Params & Pretraining data & $n$ & B-ACC \\
    \midrule
    EEGNet & 3k & none & n.r. & 0.4482 $\pm$ .0094 \\
    EEGConformer & 550k & none & n.r. & 0.4696 $\pm$ .0106 \\
    SPaRCNet & 790k & none & n.r. & 0.4635 $\pm$ .0117 \\
    ContraWR & 1.6M & none & n.r. & 0.4678 $\pm$ .0125 \\
    CNN-Transformer & 3.2M & none & n.r. & 0.4600 $\pm$ .0108 \\
    FFCL & 2.4M & none & n.r. & 0.4470 $\pm$ .0143 \\
    ST-Transformer & 3.5M & none & n.r. & 0.4575 $\pm$ .0145 \\
    BIOT & 3.2M & PREST${+}$SHHS & n.r. & 0.4748 $\pm$ .0093 \\
    LaBraM-Base & 5.8M & 2{,}535\,h & n.r. & 0.4869 $\pm$ .0085 \\
    CBraMod & 4.0M & ${>}9{,}000$\,h TUEG & n.r. & 0.5138 $\pm$ .0066 \\
    REVE-Base & 69M & 61{,}415\,h & n.r. & 0.6396 $\pm$ .0095 \\
    \midrule
    Ours --- paired init.\ (EA), full-FT & 113k & none & 3 & 0.6021 $\pm$ .0249 \\
    Ours --- TUEG SSL (cross-corpus), full-FT & 113k & 3{,}768\,h TUEG subset & 3 & 0.5495 $\pm$ .0213 \\
    Ours --- TUEG SSL (cross-corpus, EA), full-FT & 113k & 3{,}768\,h TUEG subset & 3 & \textbf{0.6108 $\pm$ .0089} \\
    Ours --- BCI-IV-2a SSL, full-FT & 113k & 3.2\,h BCI-IV-2a train & 3 & 0.5026 $\pm$ .0148 \\
    Ours --- BCI-IV-2a SSL (EA), full-FT & 113k & 3.2\,h BCI-IV-2a train & 3 & 0.6088 $\pm$ .0308 \\
    \bottomrule
  \end{tabular}
\end{table}

\section{Implementation and reproducibility details}
\label{app:enablement}
The companion repository is available at
\href{https://github.com/Sequel-Institute/rqf-ssl}{https://github.com/Sequel-Institute/rqf-ssl}.

\subsection{Model configurations}

\paragraph{Parameter accounting.}
RQF parameters are complex-valued, so we count each complex parameter as
two real scalars. Table~\ref{tab:params} applies this convention to the
complete downstream models used in the paper. Totals include the per-filter
time constants $\tau$ (256 scalars for encoders of width 64 and 512 for
width 128).

\begin{table}[h]
\centering
\small
\begin{tabular}{lrrrr}
\toprule
Model & Encoder & Summary/pooler & Readout & Total \\
\midrule
Speech classifier (d4/64) & 33.7k & --- & 35.6k & 69.3k \\
EEG windowed (d4/64) & 33.7k & 16.6k & 59.8k & 110k \\
ISRUC epoch-local windowed (d4/64) & 33.7k & 16.6k & 147k & 198k \\
ISRUC contextual windowed (d4/64) & 33.7k & 16.6k & 173k & 224k \\
BCI-IV-2a (d4/64) & 33.7k & 16.6k & 62.6k & 113k \\
iEEG windowed (d4/128) & 133k & 65.9k & 94.1k & 293k \\
\bottomrule
\end{tabular}
\caption{Real-scalar parameter counts for the downstream models, rounded
to three significant digits (k denotes thousands). Components and totals
are rounded independently. The
ISRUC contextual model adds two one-layer RQFs over the epoch
representations.}
\label{tab:params}
\end{table}

\paragraph{Multiscale frequency initialization.}
Deeper RQF layers are initialized at progressively lower frequencies, so
the stack begins with filters spanning several temporal scales. In the
four-layer encoders used here, the upper frequency limit is multiplied by
$1$, $1/2$, $1/4$, and $1/8$ from the first to the fourth layer.
This schedule affects only the starting values:
during training, every layer can learn frequencies throughout the common
range $[1/T,\,f_s/2]$.

\paragraph{Shared RQF configuration.}
All experiments use the same encoder core: a pure RQF stack with
zero-order-hold discretization, split-ReLU activations, and a prospective
two-tap input. The same stack is applied to every channel. It contains no
skip connections, dropout, or normalization between layers. We initialize
$\tau=2\Delta t$; it stays fixed during pretraining and in the TUAB and
BrainTreebank evaluations, while full fine-tuning on BCI-IV-2a, ISRUC, and
Speech Commands also updates it. We initialize the bandwidth parameter
$\gamma$ in $(0,1)$ and use a mixing-weight
gain of $0.5$ at initialization. Table~\ref{tab:hyperparams} gives the
dataset-specific widths, prediction geometry, and training settings for
TUAB and BrainTreebank.
Input preprocessing is dataset-specific and described below. Layer
normalization in the window pooler and downstream readout does not use
dataset-wide statistics. Dense targets are standardized separately from
a training-only calibration subset.

\paragraph{Coordinate-aware readout.}
Multichannel tasks use the spatial readout of Section~\ref{sec:method},
which allows one model to operate on different electrode montages.
Each integer-millimetre coordinate indexes three learned
$200\times32$ axis embeddings, covering positions from $-100$ to $+99$\,mm.
Their sum passes through a $32\to32\to16$ MLP to score 16 spatial basis
functions across the observed channels. The corresponding value vectors
have width 32, giving projections of $64\to512$ for EEG and $128\to512$
for iEEG. We concatenate the basis-weighted sums with the channelwise mean,
standard deviation, and maximum, then summarize these features over time
by their mean and standard deviation before classification. These choices
produce the readout counts in Table~\ref{tab:params}.

\begin{table}[!b]
\centering
\small
\setlength{\tabcolsep}{4pt}
\begin{tabularx}{\linewidth}{@{}>{\raggedright\arraybackslash}p{0.25\linewidth}>{\raggedright\arraybackslash}X>{\raggedright\arraybackslash}X@{}}
\toprule
 & TUAB & BrainTreebank \\
\midrule
Signal & scalp EEG & intracranial EEG \\
Encoder depth / width & 4 / 64 & 4 / 128 \\
Sampling rate $f_s$ & 128\,Hz & 2{,}048\,Hz \\
Sequence context $T$ & 100\,s & 3\,s \\
Window / step & 1\,s / 1\,s ($100$ windows) & 62.5\,ms / 62.5\,ms ($48$ windows) \\
Window summary & \multicolumn{2}{>{\raggedright\arraybackslash}p{0.72\linewidth}@{}}{complex mean $\oplus$ final state $\rightarrow$ linear + LayerNorm} \\
Summary width & 64 & 128 \\
\midrule
SSL horizon $h_w$ & 1 window (1\,s); 2 windows (2\,s) for the dense comparison & 1 window (62.5\,ms) \\
Inverse temperature $\beta$ & \multicolumn{2}{>{\raggedright\arraybackslash}p{0.72\linewidth}@{}}{10} \\
Sampled negatives & \multicolumn{2}{>{\raggedright\arraybackslash}p{0.72\linewidth}@{}}{256, sampled uniformly from other window summaries in the batch (any channel)} \\
Loss weights, global / variance-hinge & \multicolumn{2}{>{\raggedright\arraybackslash}p{0.72\linewidth}@{}}{1.0 / 0.04} \\
\midrule
Pretraining optimizer & \multicolumn{2}{>{\raggedright\arraybackslash}p{0.72\linewidth}@{}}{AdamW, lr $10^{-4}$, weight decay $5\times 10^{-2}$, grad-clip 1.0} \\
Pretraining epochs / batch & 30 / 8 & 20 / 2 \\
\midrule
Downstream input / epochs / batch & 10-s crops or 100-s segments / 6 / 128 or 16 & 3\,s / 20 / 2 (grad.\ accum.\ 2) \\
Downstream optimizer & AdamW, lr $10^{-3}$, wd $0.1$ on weight matrices, cosine, weight averaging (Section~\ref{app:training-protocols}) & AdamW, lr $2\times 10^{-4}$ pretrained and $2\times 10^{-3}$ new; warmup-exponential; wd $10^{-2}$ \\
Model selection & \multicolumn{2}{>{\raggedright\arraybackslash}p{0.72\linewidth}@{}}{validation only; test set evaluated once per configuration} \\
\bottomrule
\end{tabularx}
\caption{Windowed TUAB and BrainTreebank configurations; shared settings
span both columns. ISRUC and SC10 use separate schedules
(Section~\ref{app:training-protocols}), as does BCI-IV-2a.}
\label{tab:hyperparams}
\end{table}

\subsection{Datasets and preprocessing}
\label{app:datasets}
Published values come from their source papers and were not re-run under
our protocol. Each item below ends with the differences that remain; the
controlled comparisons share one protocol within each benchmark and are
not affected by them.
\begin{itemize}
  \item \textbf{Speech Commands (10-class)}~\cite{warden2018speech}. We retain
  the 34{,}975 exactly one-second clips and use a fixed, class-stratified
  70/15/15 split at the clip level. Waveforms are sampled at 16\,kHz and
  standardized using training-set statistics.
  \item \textbf{TUAB (EEG, abnormal detection).} We derive the TCP-16
  bipolar montage from the 19 referential 10--20 channels and resample it
  to 128\,Hz. No notch filter is applied. Recordings are divided into non-overlapping 100-second
  segments and normalized independently as described below. Following the
  BIOT convention~\cite{yang2023biot}, we divide the non-test patients
  80/20 into training and validation sets with seed 0. The held-out test
  cohort and its provenance are described in Appendix~\ref{app:additional-results}.
  We evaluate on 10-s crops of these segments and on whole 100-s segments.
  LaBraM, CBraMod, and REVE report 10-s segments at 200\,Hz; SAMBA reports
  10-s and 100-s segments at 128\,Hz.
  \item \textbf{ISRUC-SG1 (EEG, five-class sleep staging).} We use the
  NeMAR v1.0.1 release and the sleep-stage labels provided by scorer 1.
  We use six
  EEG channels (F3--A2, C3--A2, O1--A2, F4--A1, C4--A1, and O2--A1) at
  200\,Hz and score 30-second epochs~\cite{khalighi2016isruc}.
  Each recording is filtered at 0.3--35\,Hz with a fifth-order Butterworth
  filter and at 50\,Hz with a notch filter ($Q=30$), before division into
  epochs. Subjects 1--80, 81--90,
  and 91--100 define the training, validation, and test splits; I008 is
  excluded because F3 and F4 are absent from the released recording. SSL
  uses only the training subjects. For downstream evaluation, we divide
  each recording into non-overlapping blocks of 20 consecutive labeled
  epochs. Unknown-stage epochs interrupt a block, and incomplete blocks
  are discarded without padding, leaving 87{,}100 epochs. Both RQF
  configurations are evaluated on these same retained epochs; only the
  contextual model uses information from neighboring epochs. The published
  values use 80 training subjects, a different context module, and
  parameter counts that omit it; the caption of Table~\ref{tab:isruc}
  gives the details.
  \item \textbf{BCI-IV-2a (EEG, four-class motor imagery).} We retain all
  5{,}184 trials and use the 22 EEG channels, excluding EOG. Signals are
  common-average referenced and filtered with a causal fifth-order
  Butterworth bandpass at 0.3--40\,Hz. We extract the four-second cue
  interval and resample it from 250 to 200\,Hz. Subjects 1--5, 6--7, and
  8--9 provide 2{,}880 training, 1{,}152 validation, and 1{,}152 test
  trials, respectively. The downstream readout uses two non-overlapping
  two-second windows per trial. When Euclidean alignment is used, its transformation is
  estimated separately from the unlabeled trials of each recording
  session, including validation and test sessions, and applied to all
  trials in that session. In-domain SSL sees only training-subject trials.
  We transfer only the encoder; the downstream learned window pooler and
  spatial readout are initialized anew. The published values come from
  REVE Table~12 with REVE's released split; REVE applies Euclidean
  alignment, and the preprocessing and repeat counts of the other models
  are not reported.
  \item \textbf{BrainTreebank (iEEG, sentence onset).} We use three-second
  segments at 2{,}048\,Hz, retaining the channels specified by the
  clean-channel metadata and excluding contacts labeled GRID or VENT.
  Signals are Laplacian-referenced using adjacent contacts and
  notch-filtered at 60\,Hz and its harmonics through 360\,Hz ($Q=30$).
  Each segment is standardized separately within each channel.
  Electrode coordinates are provided to the downstream spatial readout.
  We use BaRISTA's chronological fold generator: seven held-out sessions
  with five folds each. The published values in BaRISTA's Appendix
  Table~16 use the same sessions and folds and report mean$\pm$s.e.m.;
  they are not paired with our results.
\end{itemize}

\paragraph{Segment normalization.}
To reduce sensitivity to isolated amplitude extremes, TUAB segments are
scaled independently in each channel:
$x \mapsto x / \bigl(q_{95}(|x|) + 10^{-8}\bigr)$, where the percentile is
computed over time. This operation neither centers nor clips the signal.

\subsection{Pretext construction}

\paragraph{Pretext-task geometry.}
The two objectives use different units of prediction. Windowed prediction
groups the encoder states into non-overlapping intervals, whereas dense
prediction retains a source at every timestep. In the dense case,
$N_w$ in Eq.~\eqref{eq:dense_logpower} controls only the interval used to
measure future power; it does not pool the source representation.

For a window with sample indices $\mathcal{I}_j$, the mean summary is
\[
\bar{\mathbf{t}}_j = \frac{1}{|\mathcal{I}_j|}
\sum_{n \in \mathcal{I}_j}
[\mathrm{Re}\,\by_n^{(L)} \,;\, \mathrm{Im}\,\by_n^{(L)}].
\]
Writing $m_j = \max \mathcal{I}_j$ and
$\mathbf{s}_j = [\mathrm{Re}\,\by_{m_j}^{(L)} \,;\,
\mathrm{Im}\,\by_{m_j}^{(L)}]$ for the final state in the window, the
learned pooler produces
\[
\mathbf{t}_j = \mathrm{LN}\!\left(
\mathbf{P}[\bar{\mathbf{t}}_j \,;\, \mathbf{s}_j]\right).
\]
The parameter-free configuration uses
$\mathbf{t}_j = \bar{\mathbf{t}}_j$ directly.

\begin{table}[h]
\centering
\small
\setlength{\tabcolsep}{3.5pt}
\begin{tabular}{@{}llllll@{}}
\toprule
Family & Input $[T \times C]$; rate & Encoder & Windows & Summary & Horizon \\
\midrule
Speech dense & $16{,}000 \times 1$; 16\,kHz & d4/64 & --- ($N_w{=}400$) & 128/64 & 1{,}600 smp $=$ 100\,ms \\
Speech windowed (mean) & $16{,}000 \times 1$; 16\,kHz & d4/64 & $40 \times 400$ & 128 & 4 win. $=$ 100\,ms \\
EEG windowed+pooler & $12{,}800 \times 16$; 128\,Hz & d4/64 & $100 \times 128$ & 64 & 2 win. $=$ 2{,}000\,ms \\
EEG dense & $12{,}800 \times 16$; 128\,Hz & d4/64 & --- ($N_w{=}128$) & 128/64 & 256 smp $=$ 2{,}000\,ms \\
iEEG windowed+pooler & $6{,}144 \times C$; 2{,}048\,Hz & d4/128 & $48 \times 128$ & 128 & 1 win. $=$ 62.5\,ms \\
\bottomrule
\end{tabular}
\caption{Pretext geometry used for each family. Windowed horizons are in
windows ($h_w$), dense horizons in samples ($\Delta$). ``Summary'' gives
the dimensionality entering the prediction head (dense: predictor source /
target key). The executed EEG windowed-vs-dense pair is horizon-matched at
2\,s (2 windows vs 256 samples).}
\label{tab:pretext_geometry}
\end{table}

\paragraph{Contrastive objective parameters.}
Every contrastive configuration uses cosine similarity with inverse
temperature $\beta=10$, 256 sampled negatives, and detached targets.
Windowed prediction on TUAB, BrainTreebank, ISRUC, and SC10 samples other
window summaries from the batch with replacement. BCI-IV-2a and its TUEG
pretraining source instead sample 256 distinct summaries from the same
channel in other examples. The source and positive windows themselves
never overlap. Single-channel dense prediction excludes
nearby anchors whose target intervals overlap, while multichannel dense
prediction draws from the same channel in other examples.

Dense targets are standardized coordinate by coordinate using 256 seeded
training examples, with a standard-deviation floor of $10^{-6}$ and
$\epsilon=10^{-6}$ in Eq.~\eqref{eq:dense_logpower}. The windowed objective
adds a variance-hinge penalty of weight $0.04$ to discourage collapsed
summaries; the dense objective uses only the contrastive loss.

\paragraph{Prediction-horizon selection.}
The horizon should be long enough that copying the present is ineffective,
but short enough that the future remains predictable. For the matched
TUAB windowed--dense comparison, a preregistered diagnostic selects the shortest candidate for
which a persistence predictor has
copy-$R^2\leq0.50$ while a ridge-affine predictor retains
$R^2\geq0.20$. On the 100-second recordings, copy-$R^2$ falls from $0.99$
at 125\,ms to $0.14$ at 2\,s, while affine-$R^2$ remains $0.70$ at 2\,s;
the rule therefore selects 2\,s. The headline TUAB configuration retains
the earlier one-second horizon. A validation-only comparison on ISRUC
likewise selects two one-second windows over one.

The 100-ms speech horizon predates this rule. A later controlled screen
found that shortening it to 12.5\,ms made the target largely copyable and
reduced frozen-encoder accuracy by $7$--$8$ points, supporting the same
choice qualitatively.

\subsection{Training and model selection}
\label{app:training-protocols}

\paragraph{Pretraining schedules.}
TUAB and BrainTreebank use AdamW with learning rate $10^{-4}$, weight
decay $5\times10^{-2}$, and gradient clipping at $1.0$; SC10 uses learning
rate $10^{-3}$ and weight decay $10^{-2}$.
BCI-IV-2a and TUEG pretraining use AdamW with peak learning rate $10^{-4}$
and weight decay $0.05$. The learning rate warms up for five and two
epochs, respectively, then follows a cosine schedule ending at 5\% of
its peak. Weight decay applies to all
optimized parameters, including the frequency and
bandwidth parameters $(\bomega,\bgamma)$.
Downstream
evaluation always starts from the final pretraining epoch, so validation
performance never selects the SSL checkpoint.
ISRUC pretraining runs for 30 epochs on individual 30-second training
epochs, using one-second windows and a two-second horizon. SC10 pretraining
runs for 50 epochs. For BCI-IV-2a, in-domain SSL runs for 60 epochs using
eight non-overlapping half-second windows per trial and a one-second
prediction horizon; the cross-corpus models use the final TUEG checkpoint, trained
for 30 epochs on a 3{,}768-hour subset.

\paragraph{Downstream training.}
Encoder-only transfer reuses the pretrained encoder and initializes the
downstream pooler and readout anew. Encoder-and-pooler transfer also
reuses the pretrained window pooler. The SSL prediction head is
discarded in both cases.

The default fine-tuning settings use AdamW with weight decay $10^{-2}$ on all
optimized parameters, including $(\bomega,\bgamma)$ when trained. Pretrained
parameters receive learning rate $2\times10^{-4}$ and newly initialized
readout parameters receive $2\times10^{-3}$; both rates warm up linearly
for five epochs and then decay exponentially. The frozen BrainTreebank
windowed evaluations keep both the encoder and transferred pooler fixed,
optimizing only the spatial readout and classifier at the latter rate.
TUAB and ISRUC fine-tune pretrained and paired-initialization models with
shared fine-tuning settings: AdamW at learning rate $10^{-3}$ with weight decay on
weight matrices only ($0.1$ for TUAB, $0.05$ for ISRUC), one warmup epoch
followed by cosine decay, gradient clipping at $1.0$, an exponential
moving average of the weights, and channel dropout and gain augmentation.
The averaged weights with the best validation AUROC (TUAB) or Cohen's
$\kappa$ (ISRUC) are selected. TUAB trains for six epochs on random 10-s
crops of each 100-s segment, or on whole segments for 100-s inputs; frozen
TUAB models train only the spatial readout and classifier.
ISRUC transfers the encoder alone and trains the pooler, spatial readout,
context RQFs, and classifier for 30 epochs with class-weighted
cross-entropy. The epoch-local ISRUC comparison and BCI-IV-2a use the
default settings above with a 50-epoch cap. SC10 follows its separate community settings for up to 200
epochs: AdamW at $10^{-2}$ with no weight decay and $\epsilon=10^{-4}$.

\paragraph{Reporting and model selection.}
Unless stated otherwise, results are reported as mean $\pm$ standard
deviation over three downstream training runs with different random
seeds, all starting from one fixed checkpoint. This measures variation
in downstream training, conditional on that checkpoint. We select the
downstream epoch using validation data and evaluate the test set once at
that epoch.

\paragraph{Low-label evaluation.}
At label budgets of 1, 5, 10, and 25 percent, we construct nested TUAB
patient subsets using sampling seed 0, retaining both classes at every
budget. The 1\% condition contains 16 training patients and 279 100-s
segments.
Larger budgets retain the previously selected patients; validation and
test patients remain unchanged. Each pretrained encoder is compared with its
byte-identical initialization before SSL, under both frozen and full
fine-tuning, with three downstream seeds per condition.

\section{Related work}
\label{app:related}

\subsection{Predictive representation learning}

Contrastive predictive coding (CPC) predicts future latent
representations from an autoregressive context~\cite{oord2018cpc}.
Later speech methods instead mask the latent sequence: wav2vec~2.0
contrasts quantized targets, while HuBERT predicts clustered
targets~\cite{baevski2020wav2vec,hsu2021hubert}. Joint-embedding
predictive architectures predict representations of masked image or
video regions using an exponential-moving-average target
encoder~\cite{assran2023ijepa,bardes2024vjepa}.

Our prediction axis is time along the same trajectory. The windowed
InfoNCE configuration is closest to CPC. Its regression variant shares
the feature-prediction form used by JEPA, but predicts a later state of
the same encoder instead of using a separate target encoder. Neither
configuration masks the input or requires a quantizer.

\subsection{Self-supervision with state-space encoders}

State-space encoders have also been used with several self-supervised
objectives. Audio Mamba reconstructs masked spectrogram patches with
input-dependent state transitions and slightly fewer parameters than its
matched Transformer~\cite{yadav2024audiomamba,gong2022ssast}. RQF poles are
instead input-independent, and the SC10 windows summarize filter states
computed directly from the waveform. Related work includes CPC with a
structured SSM for ECG, contrastive S4 variants, and predictive,
joint-embedding, diffusion, and state-retention objectives for ECG and
EEG~\cite{mehari2023quantitative,almasud2026pretraining,wu2026sls4wave,
hong2025samba,broustail2026lumamba,puah2025eegdm,durgam2026cambrain}.
Our scope is the future-state objective for a compact complex-valued RQF;
predictive self-supervision with state-space encoders is already established.

\subsection{EEG and intracranial foundation models}

The EEG references in Figure~\ref{fig:pareto} span several pretraining
objectives. LaBraM, CBraMod, and REVE use masked prediction with Transformer
encoders; BIOT learns contrastively across datasets; and SAMBA uses masked
reconstruction with a Mamba encoder--decoder~\cite{jiang2024labram,
wang2025cbramod,reve2025,yang2023biot,hong2025samba}. Their TUAB
configurations range from 1M to 369M parameters.

For intracranial recordings, Brant, BrainBERT with a Population Transformer,
and BaRISTA provide the 1M--500M parameter references~\cite{zhang2023brant,
wang2023brainbert,chau2025popt,oganesian2025barista}. Figure~\ref{fig:pareto}
uses BaRISTA's chronological evaluation, with the same sessions and folds as ours.

\end{document}